\documentclass{article}
\pdfoutput=1
\PassOptionsToPackage{square,sort,comma,numbers}{natbib}

\usepackage{adjustbox}

 \usepackage[final]{neurips_2022}

\usepackage[utf8]{inputenc} % allow utf-8 input
\usepackage[T1]{fontenc}    % use 8-bit T1 fonts
\usepackage{hyperref}       % hyperlinks
\usepackage{url}            % simple URL typesetting
\usepackage{booktabs}       % professional-quality tables
\usepackage{amsfonts}       % blackboard math symbols
\usepackage{nicefrac}       % compact symbols for 1/2, etc.
\usepackage{microtype}      % microtypography

\usepackage{multicol}
\usepackage{multirow}

\usepackage{makecell}

\usepackage{tabularx}

\usepackage{tabularx}

\usepackage{xcolor}
\usepackage[linesnumbered,ruled,vlined]{algorithm2e}

\title{Another algorithm template}
\author{Roy}

\SetCommentSty{mycommfont}

\usepackage{booktabs} 

\usepackage[linesnumbered,ruled,vlined]{algorithm2e}

\usepackage[T1]{fontenc}
\usepackage[utf8]{inputenc}
\usepackage{babel}
\usepackage[font=small,labelfont=bf]{caption}
\usepackage{subcaption}
\usepackage{graphicx}
\usepackage{capt-of}% or \usepackage{caption}
\usepackage{xcolor,colortbl}
\usepackage{footmisc}
\usepackage{soul}

\definecolor{LightCyan}{rgb}{0.88,1,1}

\usepackage[utf8]{inputenc} % allow utf-8 input
\usepackage[T1]{fontenc}    % use 8-bit T1 fonts
\usepackage{hyperref}       % hyperlinks
\usepackage{url}            % simple URL typesetting
\usepackage{booktabs}       % professional-quality tables
\usepackage{amsfonts}       % blackboard math symbols
\usepackage{nicefrac}       % compact symbols for 1/2, etc.
\usepackage{microtype}      % microtypography
\usepackage{xcolor}         % colors

\title{Neural Routing in Meta Learning}

\author{%
Jicang Cai$^*$ \quad Saeed Vahidian$^*$ \quad
\textbf{Weijia Wang$^*$}
\quad \textbf{Mohsen Joneidi} \quad \textbf{Bill Lin} \\\\
University of California San Diego 
}

\begin{document}

\maketitle

\begin{abstract}
Meta-learning often referred to as learning-to-learn is a promising notion raised to mimic human learning by exploiting the knowledge of prior tasks but being able to adapt quickly to novel tasks. A plethora of models has emerged in this context and improved the learning efficiency, robustness, etc. The question that arises here is can we emulate other aspects of human learning and incorporate them into the existing meta learning algorithms? Inspired by the widely recognized finding in neuroscience that distinct parts of the brain are highly specialized for different types of tasks, we aim to improve the model performance of the current meta learning algorithms by selectively using only parts of the model conditioned on the input tasks. In this work, we describe an approach that investigates task-dependent dynamic neuron selection in deep convolutional neural networks (CNNs) by leveraging the scaling factor in the batch normalization (BN) layer associated with each convolutional layer. The problem is intriguing because the idea of helping different parts of the model to learn from different types of tasks may help us train better filters in CNNs, and improve the model generalization performance. We find that the proposed approach, neural routing in meta learning (NRML), outperforms one of the well-known existing meta learning baselines on few-shot classification tasks on the most widely used benchmark datasets.
\end{abstract}

\section{Introduction}

Few-shot classification or learning a classifier to generalize to unseen classes by using a limited number of labeled data has attracted remarkable attention~\cite{vinyals2016matching, koch2015siamese, chen2019closer}. Meta-learning algorithms can learn to quickly adapt to unseen tasks by extracting transferable knowledge from few examples~\cite{mishra2017simple, finn2017model, snell2017prototypical}. Broadly speaking, in the paradigm of meta-learning, algorithms can be divided into two main approaches.  The first approach, i.e., ``learning to compare" (non-parametric) tends to learn an appropriate embedding function, so that prediction is based on the distance of a new example to the labeled examples~\cite{snell2017prototypical, matchingnet2016, Ren2018, Liu2018}. The second one is ``learning to optimize" (optimization-based), which tends to develop a learning algorithm that can learn a new episode efficiently via only few steps of parameter updating~\cite{finn2017model, sachin2016, Mishra2017, reptile2018, rusu2018}. Non-parametric few-shot learning methods have the advantage that learned embedding space could be used in target task without explicitly design the architecture for the desired number of classes. However, they do not adapt the network weights to the target task. On the other hand, optimization-based algorithms have the power to adapt to an unseen task with gradient descent and can better take advantage of the provided information for new unseen task training~\cite{Vahidian-ICLR-2021}.  In comparison with different approaches such as metric-based algorithms which are more suitable for non-parametric learners, optimization-based algorithms are simpler but also more general and thus have been applied to a variety of applications. In this paper, we elaborate on  “learning to optimize” framework. This is while the other framework can also be incorporated into our model. Our proposed method allows optimization-based methods to be much more efficient in terms of memory usage since they do not need to keep the whole training path in memory and can updates selected neurons (filters in CNN).

% However, they usually require a lot of memory and cannot be trained on tasks with large training data or which require too many inner loop iterations.

In this paper, we rely on recent advances in the field of human brain/memory which is often referred as an informational processing system. It plays the pivotal role in human intelligence and has inspired many well-known machine learning models.  It is broadly recognized in neuroscience that different parts of the brain are highly specialized for distinct types of tasks~\cite{brain}. It contributes not only to the high efficiency in handling a response but also the surprising efficacy of the brain in learning novel tasks.

Episodic memory of brain, as a longterm memory, is the collection of past human experiences. They can be retrieved and exploited by the brain when tackling problems that have never been occurred before. Different memories activate different neurons in the brain, directing us to perform well on what we have not done before. Inspired by the above-mentioned observations in neuroscience we propose why not emulating this learning process to the existing meta/few-shot learning algorithms which strive for reducing the gap between human learning and machine leaning models. More specifically, we describe Neural Routing in Meta learning (NRML), as novel method that learns each speciﬁc task, only by involving a small portion of the model in a meta learning algorithm. The small portion of the neurons are selected by leveraging the scaling factor $\gamma$ in batch normalization (BN) layers associated with each convolutional layer as indicators of  importance of the corresponding
neurons. It  means that only a small fraction of neurons needs to be updated at each back-propagation step, which is desirable for training a large model, facilitates the difficulty of learning, and achieves better performance.

% this observation, we propose a new the routing-by-neuron mechanism in meta-learning paradigm. Inspired by the the above-mentioned characteristics, we describe a similar mechanism in  meta-learning models. 

% based on just the  value of parameters in batch normalization in inner and outer loops optimization. The main idea of these papers is that we can leverage...

% Meta-learning algorithms can learn to quickly adapt to unseen tasks by extracting transferable knowledge from few examples~\cite{mishra2017simple, finn2017model, snell2017prototypical}.  ning algorithms can learn to quickly adapt to unseen tasks by extracting transferable knowledge from few examples~\cite{mishra2017simple, finn2017model, snell2017prototypical}. 

% [optimization-based] However, they usually require a lot of memory and cannot be trained on tasks with large training data or which require too many inner loop iterations. There has been recent novel approaches in both types of algorithms that address these issues. \cite{rajeswaran2019meta} proposed iMAML, which leverages implicit gradient and calculates the update rule based on just final parameters of the task and by using an iterative solver. This allows optimization-based methods to be much more efficient in terms of memory usage since they do not need to keep the whole training path in memory and can estimate updates based on just the final value of parameters after inner loop optimization. 

\section{Related Work}\label{related-work}
The approach that we propose in this paper addresses the  meta-learning for classification which aims to obtain transferable knowledge from a few examples~\cite{hsu2018unsupervised, khodadadeh2019unsupervised, AAL2019} and from a few neurons. In the following paragraphs in this section, we describe the prior work in meta/few-shot learning,  and Sub-Network Routing, as they are the most related topics to this work.

{\bf{Meta/Few-Shot Learning.}} Few-shot classification aims at learning a model that can be efficiently adapted to  unseen classes from few samples. Early methods~\cite{vinyals2016matching, sachin2016, finn2017model, Mishra2017, rusu2018, reptile2018, Ren2018, AAL2019, rajeswaran2019meta} pose the few-shot classification problem in a learning-to-learn paradigm by training a deep network over a distribution of related tasks which are constructed from the support set, and transfer this experience to enhance its performance for learning novel and unseen classes. Concretely,~\cite{vinyals2016matching} learn a feature encoder that
is conditioned on the training set in meta-training and does
not necessitate any further training during meta-test due to its nonparametric classifier. The authors in~\cite{sachin2016} leverage the idea of learning a feature encoder in meta-train further and learning an update rule via an LSTM to 
update the classifier in meta-test. \cite{finn2017model} poses the
problem as a meta-learning formulation and learn the parameters of
a deep network in meta-training such that a neural network initialized with the learned parameters can be quickly adapted to unseen tasks. We refer to~\cite{survey-meta-2022-Timothy,survey-meta-2020-wang} for comprehensive review of early works.

{\bf{Routing on Neural Network.}} Routing on deep neural networks refers to activating only some of the modules in a network during training and inference~\cite{Neural-Routing-2021, Vahidian-federated-2021, sara-sabour-capsule}. Recent researches promoted it on CNNs in order to accelerate network inference. In AIG~\cite{Andreas-routing-2020}, BlockDrop \cite{BlockDrop}, and SkipNet~\cite{SkipNet} a subset of needed blocks is learned to process a given task. Since deep layers’ features may not be required for classification, SACT~\cite{SACT-2016}, Inside Cascaded CNN~\cite{CNN-2017}, and Dynamic Routing~\cite{dynamic-routing-2017, PathNet-2017} suggest to do input-dependent early stopping at inference time. Routing Convolutional Network (RCN)~\cite{anytime-routing-2021} introduces a routing network aimed at reducing the models’ computational costs at the cost of losing accuracy performance to some extent. Another line of work that is closely related to routing is Mixture of Experts (MoE)~\cite{MOE-hinton-2017}, where several sub-networks are exploited through an ensemble using weights determined by a gating module. From other stream of works that resemble routing on neural networks is dynamic network configuration~\cite{Bengio-2015-routing, Zhourong-2019-routing}. In dynamic network configuration, the neurons, layers or other components of the model for each input task is  dynamically selected. In these method, one small sub-module is added to each position to be conﬁgured in the model.

Some common weakness of these routing approaches are the following: i) They require an extra module (the gater network) that needs to be trained jointly with the backbone in an end-to-end fashion through back-propagation. This consumes additional memory and storage ii) Since they require some parallel modules along with the main network, the inference efficiency is lower than that of the baseline. iii) The model training is unstable and they experience accuracy drop while requiring orders of magnitude more labeled training data.

Our proposed approach, NRML, takes the advantage of meta/few-shot learning along with the selectivity knowledge of neurons to find implicit routing on the neural network. The neurons are selected based on their sensitivity (how strongly they are fired) to each task in meta training and validation stage. NRML is simple to implement and can be added to any few-shot learning algorithm. Further, it does not require any gater network to be trained along with the main network and it does not require extra memory. More importantly, it improves the accuracy performance of the meta learning baselines on the downstream unseen tasks.

\section{Method}

In this section, we provide required preliminaries and describe how our approach inspired from human learning improves the model performance of the current
meta learning algorithms by selectively using only parts of the model based
on the input tasks in meta training and validation.

\section{Preliminaries}\label{prelim}

\noindent \textbf{Batch normalization (BN)}. Let $x_{in}$ and $x_{out}$ be the input and output of a BN layer, $\mathcal{B}$ denotes the current minibatch, BN layer do the following transformation: $\hat x =\frac{x_{in}-\mu_{\mathcal{B}}}{\sqrt{\sigma_{\mathcal{B}}^2+\epsilon}}$; $x_{out}=\gamma \hat x+\beta$, where $\mu_{\mathcal{B}}$ and $\sigma_{\mathcal{B}}^2$ are the mean and standard deviation of input activations over $\mathcal{B}$; $\gamma$, and $\beta$ are trainable scale and shift parameters.

\noindent \textbf{Generating meta-tasks}. Concretely, we are dealing with an $N$-way, $(K^{(tr)}, K^{(val)})$-shot supervised learning task, $\mathcal{T}_{1}, \mathcal{T}_{2},..., \mathcal{T}_{n}$ that are drawn from an underlying joint distribution ${\cal{P^T_{X,Y}}}(x_i, y_i)$. Each task consists of two disjoint sets $D_{\mathcal{T}}^{(tr)}$ and $D_{\mathcal{T}}^{(val)}$.
$D_{\mathcal{T}}^{(tr)}$, has $i \in \{1, \ldots, N \times K^{(tr)}\}$ data points $(x_i, y_i)$ such that there are exactly $K^{(tr)}$ samples for each categorical label $y_i \in \{1, \ldots, N\}$. $D_{\mathcal{T}}^{(val)}$ contains another $N \times K^{(val)}$ data points separate from the ones in $D_{\mathcal{T}}^{(tr)}$. We have exactly $K^{(val)}$ samples for each class in $D_{\mathcal{T}}^{(val)}$ as well.  Our goal is to learn a classifier $f_{\theta}(x)$ to predict $y$ given $x$. In other words, $f_{\theta}(x) = {\rm{arg\,max}}_{y}{{\cal{P^T_{Y|X}}}(y|x)}$. We do not have access to the underlying distribution ${\cal{P^T_{X,Y}}}(x, y)$, we rather have access to a few samples of the task train set, $D_{\mathcal{T}}^{(tr)}$. 

% We can train a neural network, $f_{\theta}(x)$, on the task train set.

\subsection{Neuron Selection in Meta-Tasks}

\begin{figure}[htbp]
    \centering     
   \includegraphics[width=1\columnwidth,height=6cm]{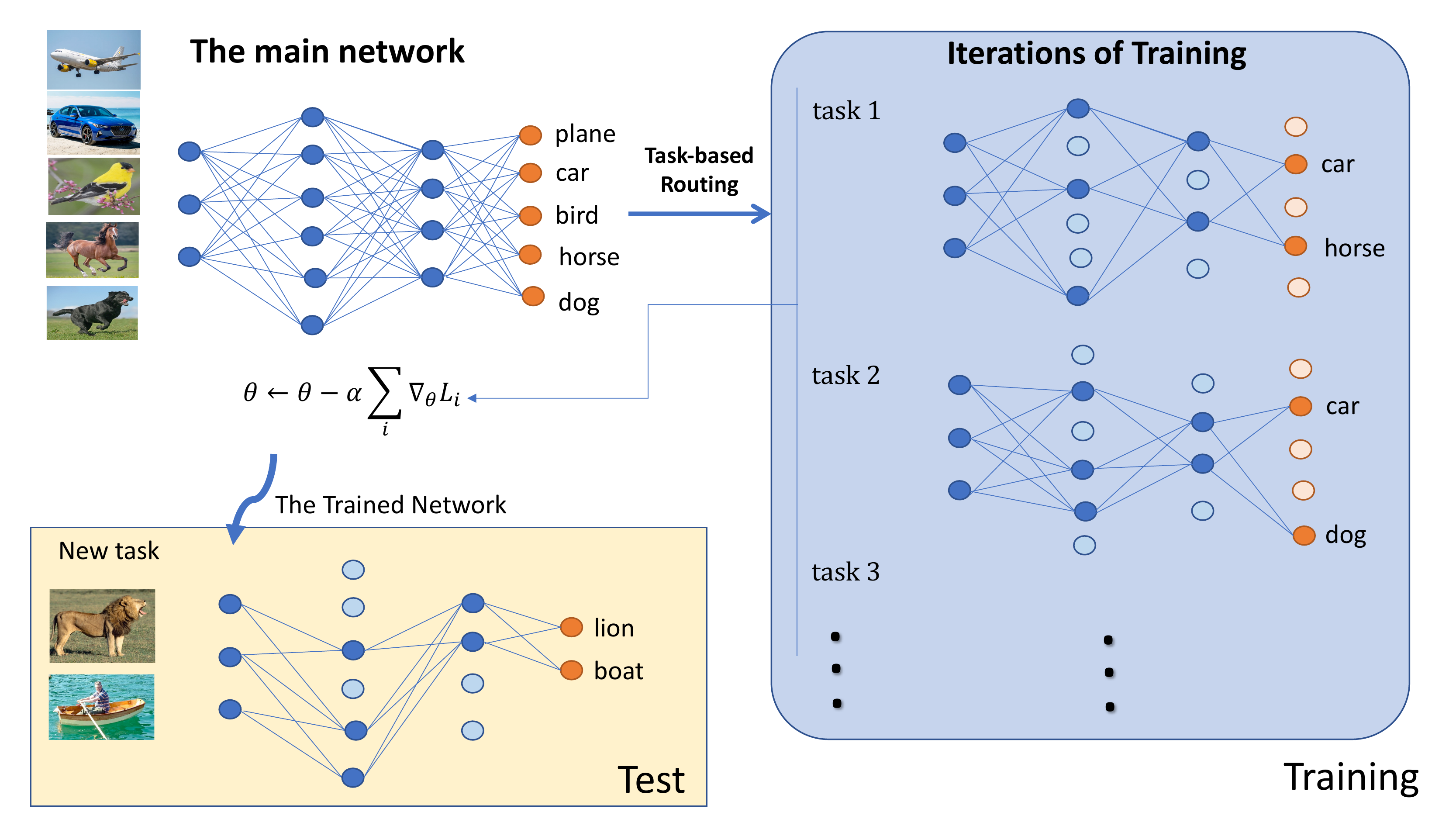}
\caption{The overview of NRML.}
    \label{fig:model}
    \vspace{-5mm}
\end{figure}

While machine learning setups have excelled humans at many tasks, they generally need far more data to achieve similar performance. Human brain can recognize new object categories based on a few instance images. It is not entirely fair to compare humans to algorithms learning from scratch, because humans learn the task with a huge amount of prior information, encoded in their brains and DNA. Rather than learning from scratch, they are recombining a set of encoded skills in their neurons. To emulate human learning, we exploit the meta learning paradigm in a novel way to encode specific knowledge in distinct neurons of the neural network which  specialized for. In our NRML, we describe a simple yet effective mechanism, an input-dependent dynamic neuron selection in convolutional neural networks (CNNs) in a few-shot learning paradigm as shown in Fig.~\ref{fig:model}. We adopt a custom network architecture meta-learning model which consists of four stacked convolutional blocks with a BN layer after each convolutional layer, with neuron-wise scaling/shifting parameters. Owing to the fact that each neuron extracts a certain pattern related to the task at hand~\cite{Like-What-You-Like}, we make use of the selectivity knowledge of neurons during training and validation.

Herein, we can directly leverage the $\gamma$ parameters in BN layers as the scaling factors we need for neuron selection. In effect, our approach introduces a scaling factor $\gamma$ for each neuron (channel), which is multiplied to the output of that neuron which facilitate its implementation without introducing any modification to existing CNN architectures. To begin, we consider a model represented by a parametrized function $f_\theta$ with parameters $\theta$. We sample a batch of tasks as described in Section~\ref{prelim}. Each task is fed to the network in the inner loop and we train these associated scaling factors. We then backpropagate the gradient only to these scaling factors $\gamma$s. The neurons in each layer is sorted based on the value of their scaling factor values. Finally we update the neurons whose corresponding $\gamma$, is among top $p\%$. The rest of neurons will not be updated for that particular task. When adapting to a task $\mathcal{T}_i$, the selected neuron’s parameters $\theta$ become $\theta'_i=\theta -\eta \nabla_\theta {\mathcal{L}}_{{\mathcal{T}}_i}(f_\theta)$ with $\eta$ being the step size. 

After feeding each task in each batch in the inner loop, we feed the validation data and the loss across all tasks within each batch is accumulated to train the model parameters in the outer loop by optimizing for the performance of $f_{\theta'_{i}}$ with respect to $\theta$ across tasks. The expected meta-objective is defined as 
\begin{equation}
{\rm{min}}_{\theta}  \sum_{\mathcal{T}_i} {\mathcal{L}}_{{\mathcal{T}}_i}(f_{\theta'_i}) = \sum_{\mathcal{T}_i} {\mathcal{L}}_{{\mathcal{T}}_i}(f_{\theta -\eta \nabla_\theta {\mathcal{L}}_{{\mathcal{T}}_i}(f_\theta)})
\label{eq1}
% \vspace{-5mm}
\end{equation}
% \vspace{-5mm}
Then, the same process is done as in inner loop to select that neurons that are strongly fired. In particular, the gradient of loss in Eq.~\ref{eq1} is backpropagated to the scaling factors of the BN layers and the top $p\%$ of them are selected whose corresponding neuron parameters are updated as 
\begin{equation}
\theta \leftarrow {\theta-\alpha \nabla_\theta \sum_{\mathcal{T}_i} {\mathcal{L}}_{{\mathcal{T}}_i}(f_{\theta'_i}) }
% \vspace{-5mm}
\end{equation} 
\noindent where $\alpha$ is the meta step size. The goal is based on the simple intuition that if a neuron is activated in certain task, that implies this neuron is able to  better extract properties that may relate to the task. Such encoded knowledge in each neuron is valuable for the network since it provides an explanation to the final prediction of the model. As a result, we propose to strengthen the weights of selected neurons. We also tried selecting neurons based on the absolute value of the activation function of each neuron for each task. However, we found that the scaling factor of BN better captures what kind of inputs can fire each neuron.

% The average estimate of the expected error on a mini-batch of tasks drawn from $\{{\mathcal{T}}_{i}\}$ is what we aim to minimize during meta-learning. 

Once the meta-learning algorithm is applied, we evaluate our meta-learned model on a set of tasks $\{{\mathcal{T}'}_{i}\}$ which do not share any instance and even classes with the tasks from $\{{\mathcal{T}}_1,...,{\mathcal{T}}_n\}$ to evaluate the capability to adapt to new unseen tasks. The pseudo-code of the NRML algorithm is described in Algorithm~\ref{alg:1}.

\SetKwComment{Comment}{/* }{ */}
\begin{algorithm}
\caption{The NRML Algorithm.}\label{alg:1}
\SetKwInOut{Require}{require}

\Require{Labeled dataset 
$\mathcal{L}=\{x_i, y_i\}$ }

\Require{$N$: class-count, $N_\mathit{MB}$: meta-batch size} 

\Require{$\alpha ,\eta$: step size hyperparameters}

randomly initialize $\theta$ \;
 \While{not done}{

\For {i in $1, \ldots, N_\mathit{MB}$} {

Sample N class from the dataset and K-shot from each class for task ${\cal{T}}_i$ for meta-train: $D_i=\{(x_1,y_1),...,(x_N,y_N)\}$\;

\For {\textbf{each} ${\cal{T}}_i$}{

Feed ${\cal{T}}_i$ to the neural net\;

 Evaluate ${\nabla _{\theta }}{{\cal{L}}_{{{\cal{T}}_i}}}\left( {{f_{\theta }}} \right)$\;
 
Backprop the loss and update only the parameters of the batch norm layer, ${BN}_{\gamma, \beta}(x_i)$\;

Store the gradient of the rest of parameters\;

Select the filters with big scaling factor values (top $k\%$), $\Omega\leftarrow$ index of the selected filters\;

Update the parameters of the filters indexed in $\Omega$ by the stored gradient and freeze the rest\;

Generate a validation set for ${{{\mathcal{T}}_i}}$: ${D_i}'=\left\{   \left ( {x_1}', y_1 \right ),..., \left ( {x_n}', y_n \right )  \right\}$ for meta-validation\;

}

Update $\theta  =\theta  -\alpha \sum\nolimits_{{{\cal{T}}_i}} {\nabla _{\theta }}{\mathcal{L}}_{\mathcal{T}_i}\left ( f_{{\theta}'_i} \right )$ using meta-validation data~${D_i}'$\;
}
 }
 
\textbf{return} $\theta$

\end{algorithm}
\vspace{-3mm}
\section{Experiments} \label{exp-main}
\vspace{-3mm}
In this section, we delve into the few-shot learning benchmark, datasets and baselines used in our evaluation as well as the implementation details. Our code is available at~\url{https://github.com/DameDollaForThree/NRML}.

\vspace{-3mm}
\subsection{Datasets}
\vspace{-2mm}
Two few-shot learning datasets, namely Omniglot and MiniImageNet, are used to evaluate the proposed NRML algorithm along with the standard MAML algorithm \cite{finn2017model} as the baseline.

{\bf{Omniglot}}~\cite{lake2011one} contains 1623 different classes. Each class corresponds to a character from 50 different alphabets. There are $20$ images associated with each character drawn by a different subject via Amazon's Mechanical Turk. We divide the dataset into 1200 characters for train, 100 characters for validation, and 323 characters for test. These characters are chosen randomly, however, by fixing the random seed for this random selection, we make sure that train, validation and test classes are disjoint sets.

{\bf{Mini-ImageNet}} is a dataset proposed for few-shot learning evaluation constructed from ImageNet images. In particular, Mini-ImageNet consists of 100 classes, with 600 64 $\times$ 64 images in each class. We adopted the partitioning of Mini-ImageNet, which divides the 100 classes into 64 classes for training, 16 classes for validation, and 20 classes for testing. These partitions are consistent across all sets of our experiments.

\vspace{-3mm}
\subsection{Neural Network Architectures}
Our models generally follow the same architecture as discussed in~\cite{finn2017model}. The model we employ for Omniglot comprises 4 blocks, each of which starts with a convolutional layer with $3\times3$ kernels and 64 channels, followed by a ReLU non-linearity and batch normalization. Strided convolutions are used rather than max-pooling to reduce the dimension. A single fully-connected layer is then placed after the last block as the classifier.
% The dimensionality of the last hidden layer is 64.
For Mini-ImageNet, the network has 4 blocks as well, which includes a convolutional layer with $3\times3$ kernels and 32 channels, followed by a ReLU non-linearity, batch normalization, and $2\times2$ max-pooling (the stride of convolutional layers is 1). The last layer is a fully connected layer with $N$ output nodes.

Though we use the specific architectures as discussed above in our evaluation, we note that our NRML algorithm can be applied to any existing CNNs architectures, including VGG, ResNet, and EfficientNet.

% We tested the proposed algorithms on two few-shot learning benchmarks: (a) the $5$-way, and $20$-way Omniglot, (b) the $5$-way, and $20$-way MiniImageNet.
\vspace{-2mm}
\subsection{Results}

In our experiments, the $(N, K)$ combinations of $(5,1)$, $(5,5)$, $(20,1)$, and $(20,5)$ were adopted for meta-training, where $N$ and $K$ stand for $N$-way $K$-shot learning. Then, $K=5$, and $K=15$ were used for validation. All experiments were repeated 3 times and we took an average over the 3 runs. The evaluation results are summarized in Tables \ref{tab:omni1}, \ref{tab:omni2}, \ref{tab:mini1}, and \ref{tab:mini2}. 

\begin{table}[ht]
% \caption{\textbf{Omniglot} N-way K-shot (N,K), $K^{val}=15$}

\caption{Accuracy results on the Omniglot dataset over $N$-way, $K^{tr}$-shot downstream
tasks with $K^{val}=15$ for each task. $\pm$ indicates the stardard deviation.
% $95\%$ confidence interval.
The top supervised result is reported in bold. The baseline results are from~\cite{finn2017model}.}

% \vspace{-3mm}
\centering
\resizebox{0.85\columnwidth}{!}{
\begin{tabular}{llllll}
            \toprule
$(N, K^{tr})$ & (5,1) &(5,5) &(20,1)& (20,5) \\
\midrule
% Epoch & $35,000$  & $55,000$  & $57,000$   &  $59,000$   &\\
% \midrule
Baseline  & $94.17 \pm 1.68$ & $98.47 \pm 0.14$& $85.22 \pm 0.84$ & $93.99 \pm 0.50$  \\ 
% \midrule
% Selection  & [0.7, 0.6, 0.4, 0.3] & [0.8, 0.8, 0.7, 0.7] & [0.8, 0.7, 0.6, 0.5] & [0.8, 0.8, 0.7, 0.7] \\ 
Ours  &  $\bf{95.51 \pm 0.32}$ & $\bf{98.68 \pm 0.04}$& $\bf{86.72 \pm 0.14}$ & $\bf{94.82 \pm 0.26}$\\  
\midrule

\end{tabular}
}
\label{tab:omni1}
\vspace{-5.5mm}
\end{table}

\begin{table}[ht]
\caption{Accuracy results on the Omniglot dataset over $N$-way, $K^{tr}$-shot downstream
tasks with $K^{val}=5$ for each task. $\pm$ indicates the stardard deviation.
% $95\%$ confidence interval.
The top supervised result is reported in bold. The baseline results are from~\cite{finn2017model}.}

% \vspace{-3mm}
\centering
\resizebox{.85\columnwidth}{!}{
\begin{tabular}{llllll}
            \toprule
$(N, K^{tr})$ = & (5,1) &(5,5) &(20,1)& (20,5) \\
\midrule
% Epoch & $49,000$  & $55,000$  & $56,000$   &  $60,000$   &\\
% \midrule
Baseline  & $94.86 \pm 0.12$ & $98.47 \pm 0.40$& $86.00 \pm 0.94$ & $93.98 \pm 0.10$  \\ 
% \midrule
% Selection  & [0.7, 0.6, 0.5, 0.4] & [0.8, 0.7, 0.5, 0.4] & [0.8, 0.8, 0.7, 0.7] & [0.8, 0.8, 0.7, 0.7] \\ 
Ours  & $\bf{96.00 \pm 0.00}$ & $\bf{98.71 \pm 0.05}$& $\bf{86.98 \pm 0.29}$ & $\bf{94.35 \pm 0.20}$ \\  
\midrule

\end{tabular}
}
\label{tab:omni2}
 \vspace{-5mm}
\end{table}

\begin{table}[h!]
% \caption{\textbf{MiniImageNet} N-way K-shot (N,K), $K^{val}=15$}

\caption{Accuracy results on the Mini-ImageNet dataset over $N$-way, $K^{tr}$-shot downstream
tasks with $K^{val}=15$ for each task. $\pm$ indicates the stardard deviation.
% $95\%$ confidence interval.
The top supervised result is reported in bold. The baseline results are from~\cite{finn2017model}.}

% \vspace{-3mm}
\centering
\resizebox{.85\columnwidth}{!}{
\begin{tabular}{llllll}
            \toprule
$(N, K^{tr})$ = & (5,1) &(5,5) &(20,1)& (20,5) \\
\midrule
% Epoch & $37,000$  & $32,000$  & $48,000$   &  $16,000$   &\\
% \midrule
Baseline  & $46.97 \pm 0.02$ & $62.37 \pm 0.47$& $17.88 \pm 0.41$ & $30.23 \pm 0.30$  \\ 
% \midrule
% Selection  & [0.7, 0.6, 0.3, 0.2] & [0.7, 0.6, 0.4, 0.3] & [0.8, 0.7, 0.6, 0.5] & [0.7, 0.6, 0.3, 0.2] \\ 
Ours  & $\bf{48.03 \pm 0.19}$ & $\bf{63.05 \pm 0.54}$& $\bf{18.57 \pm 0.18}$ & $\bf{30.74 \pm 0.25}$ \\  
\midrule

\end{tabular}
}
\label{tab:mini1}
 \vspace{-5mm}
\end{table}

\begin{table}[h!]
% \caption{\textbf{MiniImageNet} N-way K-shot (N,K), $K^{val}=5$}

\caption{Accuracy results on the Mini-ImageNet dataset over $N$-way, $K^{tr}$-shot downstream
tasks with $K^{val}=5$ for each task. $\pm$ indicates the stardard deviation.
% $95\%$ confidence interval.
The top supervised result is reported in bold. The baseline results are from~\cite{finn2017model}.}

\centering
\resizebox{.85\columnwidth}{!}{
\begin{tabular}{llllll}
            \toprule
$(N, K^{tr})$ = & (5,1) &(5,5) &(20,1)& (20,5) \\
\midrule
% Epoch & $52,000$  & $57,000$  & $25,000$   &  $24,000$   &\\
% \midrule
Baseline  & $46.74 \pm 0.47$ & $59.41 \pm 0.32$& $17.55 \pm 0.20$ & $30.03 \pm 0.59$  \\ 
% \midrule
% Selection  & [0.7, 0.6, 0.3, 0.2] & [0.7, 0.6, 0.3, 0.2] & [0.7, 0.6, 0.3, 0.2] & [0.7, 0.6, 0.3, 0.2] \\ 
Ours  & $\bf{48.03 \pm 0.65}$ & $\bf{60.22 \pm 0.32}$& $\bf{18.03 \pm 0.21}$ & $\bf{30.72 \pm 0.48}$ \\  
\midrule

\end{tabular}
}
\label{tab:mini2}
 \vspace{-3mm}
\end{table}

{\bf{Omniglot}} As can be seen from Tables \ref{tab:omni1} and \ref{tab:omni2}, our NRML approach consistently outperforms the MAML baseline in all the 8 cases. In particular, with $(N, K^{tr}, K^{val}) = (20, 1, 15)$, NRML achieves the highest advantage over MAML, which is 1.5\%.

{\bf{Mini-ImageNet}} Similar observation can be concluded from Tables \ref{tab:mini1} and \ref{tab:mini2} that NRML always achieves better results than MAML on the Mini-ImageNet dataset. $(N, K^{tr}, K^{val}) = (5, 1, 5)$ raises the largest difference between our approach and the baseline, which is about 1.3\%.

It can be observed that NRML tends to achieve higher improvement over MAML with relatively lower $K^{tr}$ (i.e., $K^{tr}=1$ vs. $K^{tr}=5$). This is potentially because that fewer samples introduce higher uncertainty and more noise in comparison with higher $K^{tr}$, while our routing algorithm reduces the effect of such uncertainty and improves the generalization of the network.

An intriguing feature of our NRML approach is that we have a clear understanding of how much of each layer in the network is taken up by the meta tasks during training. We also have a good indication of how many of the selected neurons of the previously learned tasks are being re-selected (reused). Throughout training, we observed that to get a better performance, by going deeper into the neural network, the number of selected neurons should decrease and vice versa. This observation is consistent with Fig. 3 of~\cite{CLNP-2019}. This is as expected, since by going deeper into the neural network layers, the neurons become more task specific. In our experiment on Omniglot and MiniImgaeNet the percentage of selected neurons in each layer are around $p$=[1st, 2nd, 3rd, 4th]=$[70\%, 60\%, 30\%, 20\%]$. We also observed that the first layer grows no new neurons after the early tasks is fed to the neural network which implies that the neurons fired and their corresponding features during the training of the early tasks are utilized and appeared sufﬁcient for the training of the subsequent tasks. This can be explained by the fact that the features learned by the neurons in lower layers are more general and thus more transferable in comparison with the features of the higher layers which are known to be specific. 

% ~\citecolor

% \subsection{Benny's Remarks}

% \begin{enumerate}
%     \item Selection in the outer loop does not improve the result, so all the above selections are performed for the inner loop.
%     \item Lower percentages, such as $[0.7, 0.6, 0.3, 0.2]$, generally tends to work well for MiniImageNet, but not for Omniglot.
%     \item $k^{val} = 5$ tends to fluctuate a lot more than $k^{val} = 15$, but only for Omniglot, not for MiniImageNet.
%     \item For MiniImageNet, $k^{val} = 15$ tends to work better than $k^{val} = 5$
% \end{enumerate}
\vspace{-4mm}
\section{Conclusion}
\vspace{-3mm}
In this paper, we introduced a novel mechanism, NRML which incorporates another aspects of human learning in the current meta learning paradigms. In particular, inspired from how distinct parts of the brain are highly specialized for different types of tasks, we exploit the scaling factor in the BN layer associated with each convolutional layer to select the neurons that activated by certain tasks in the train and validation process of meta learning. We found that NRML outperforms state-of-the-art MAML algorithm on the Omniglot and MiniImageNet datasets. We note that NRML can be applied to all existing meta/few-shot learning baselines.

\clearpage

\bibliography{Saeed}
\bibliographystyle{unsrt}

\end{document}